\pdfoutput=1

\documentclass[11pt]{article}

\usepackage[]{emnlp2021}

\usepackage{times}
\usepackage{latexsym}
\usepackage{amsmath}
\usepackage{mathptmx}
\usepackage{mathtools}
\usepackage{amsfonts}
\usepackage{caption}
\usepackage{subcaption}
\usepackage{graphicx}

\usepackage[T1]{fontenc}

\usepackage[utf8]{inputenc}

\usepackage{microtype}

\title{All Bark and No Bite: Rogue Dimensions in Transformer Language Models Obscure Representational Quality
}

 \author{William Timkey \and  Marten van Schijndel \\
           Department of Linguistics \\
           Cornell University \\
           \texttt{\{wpt25|mv443\}@cornell.edu}}

\begin{document}
\maketitle
\begin{abstract}
Similarity measures are a vital tool for understanding how language models represent and process language. Standard representational similarity measures such as cosine similarity and Euclidean distance have been successfully used in static word embedding models to understand how words cluster in semantic space. Recently, these measures have been applied to embeddings from contextualized models such as BERT and GPT-2. In this work, we call into question the informativity of such measures for contextualized language models. We find that a small number of rogue dimensions, often just 1-3, dominate these measures. Moreover, we find a striking mismatch between the dimensions that dominate similarity measures and those which are important to the behavior of the model. We show that simple postprocessing techniques such as standardization are able to correct for rogue dimensions and reveal underlying representational quality. We argue that accounting for rogue dimensions is essential for any similarity-based analysis of contextual language models.

\end{abstract}

\section{Introduction}

By mapping words into continuous vector spaces, we can reason about human language in geometric terms. For example, the cosine similarity of pairs of word embeddings in Word2Vec \cite{NIPS2013_9aa42b31} and GloVe \cite{pennington-etal-2014-glove} shows a robust correlation with human similarity judgments, and embeddings cluster into natural semantic classes in Euclidean space \cite{baroni-etal-2014-dont, Wang_2019}. 
In recent years, static embeddings have given way to their contextual counterparts, with language models based on the transformer architecture \cite{VaswaniSPUJGKP17} such as BERT \cite{devlin-etal-2019-bert}, RoBERTa \cite{liu2020roberta}, XLNet \cite{NEURIPS2019_dc6a7e65} and GPT-2 \cite{Radford2019LanguageMA} achieving state of the art results on many language understanding tasks. Despite their success, relatively little is known about how these models represent and process language. Recent work has employed measures such as cosine similarity and Euclidean distance to contextual representations with unclear and counterintuitive results. For example, similarity/distance measures in BERT are extremely sensitive to word position, leading to inconsistent results on evaluation benchmarks \cite{mickus-etal-2020-mean, may-etal-2019-measuring}. Additionally, representational quality appears to degrade severely in later layers of each network, with the final layers of BERT, RoBERTa, GPT-2 and XLNet showing little to no correlation with the semantic similarity/relatedness judgments of humans \cite{bommasani-etal-2020-interpreting}. 

Recent work which probes the representational geometry of contextualized embedding spaces using cosine similarity has found that contextual embeddings have several counterintuitive properties \cite{ethayarajh-2019-contextual}. For example: 1) Word representations are highly \textbf{anisotropic}: randomly sampled words tend to be highly similar to one another when measured by cosine similarity. In the final layer of GPT-2 for example, any two words are almost perfectly similar. 2) Embeddings have extremely low self-similarity: In later layers of transformer-based language models, random words are almost as similar to one another as instances the same word in different contexts.

In this work, we critically examine the informativity of standard similarity/distance measures (particularly cosine similarity and Euclidean distance) in contextual embedding spaces. We find that these measures are often dominated by 1-5 dimensions across all the contextual language models we tested, regardless of the specific pretraining objective. It is this small subset of dimensions which drive anisotropy, low self-similarity, and the apparent drop in representational quality in later layers. These dimensions, which we refer to as \textbf{rogue dimensions} are centered far from the origin and have disproportionately high variance. The presence of rogue dimensions can cause cosine similarity and Euclidean distance to rely on less than 1\% of the embedding space. Moreover, we find that the rogue dimensions which dominate cosine similarity do not likewise dominate model behavior, and show a strong correlation with absolute position and punctuation.

Finally, we show that these dimensions can be accounted for using a trivially simple transformation of the embedding space: standardization. Once applied, cosine similarity more closely reflects human word similarity judgments, and we see that representational quality is preserved across all layers rather than degrading/becoming task-specific.
Taken together, we argue that accounting for rogue dimensions is essential when evaluating representational similarity in transformer language models.%
\footnote{Our code is publically released at: \url{http://github.com/wtimkey/rogue-dimensions}}

\section{Background}

Standard measures such as cosine similarity or Euclidean distance in contextual embedding spaces have been used in a wide range of applications: to understand how the representational similarity of word embedding spaces corresponds to human semantic similarity/relatedness judgments \cite{bommasani-etal-2020-interpreting, vulic-etal-2020-probing, chronis-erk-2020-bishop, a-rodriguez-merlo-2020-word}, human brain activation patterns/cross-model similarity \cite{abnar-etal-2019-blackbox}, syntax structure \cite{chrupala-alishahi-2019-correlating}, semantic shift \cite{martinc-etal-2020-leveraging}, compositionality/idomaticity of word vectors \cite{garcia-etal-2021-probing}, polysemy \cite{soler2021lets}, context-sensitivity \cite{NEURIPS2019_159c1ffe}, social bias \cite{may-etal-2019-measuring, bommasani-etal-2020-interpreting}, changes to the embedding space during fine-tuning \cite{merchant-etal-2020-happens}, and as an evaluation metric for text generation \cite{Zhang*2020BERTScore:}.

However, a number of works have questioned the appropriateness of cosine similarity. \citet{schnabel-etal-2015-evaluation} found that static embedding models encode a substantial degree of word frequency information, which leads to a frequency bias in cosine similarity. \citet{may-etal-2019-measuring} questioned the adequacy of cosine similarity in sentence encoders after finding contextual discrepancies in bias measures. Perhaps most relevant to the present work is \citet{zhelezniak-etal-2019-correlation} which treats individual word embeddings as statistical samples, shows the equivalence of cosine similarity and Pearson correlation, and notes that Pearson correlation (and therefore cosine similarity) is highly sensitive to outlier dimensions. They further suggest the use of non-parametric rank correlation measures such as Spearman's $\rho$, which is robust to outliers. Our work investigates the sensitivity of cosine similarity to outlier dimensions in contextual models, and further characterizes the behavioral correlates of these outliers.

Our goal in this work was not causal explanation of degenerate embedding spaces or post-processing for task performance gains, but rather to empirically motivate trivially simple transformations to enable effective interpretability research with existing metrics. However, we refer interested readers to \citet{gao2018representation} who studied degeneration toward anisotropy in machine translation. Similarly, \citet{li-etal-2020-sentence} suggested a learned transformation of transformer embedding spaces which resulted in increased performance on semantic textual similarity tasks.

\section{Rogue Dimensions and Representational Geometry}
\label{sec:repgeo}
\subsection{Anisotropy}
\label{sec:2}
In this section, we investigate how each dimension of the embedding space contributes to anisotropy, defined by \citet{ethayarajh-2019-contextual} as the expected cosine similarity of randomly sampled token pairs. They showed that contextual embedding spaces are highly anisotropic, meaning that the contextual representations of any two tokens are expected to be highly similar to one another. We investigate this counterintuitive property by decomposing the cosine similarity computation by dimension, and show that the cosine similarity of any two tokens is dominated by a small subset of \textbf{rogue dimensions}. We conclude that anisotropy is not a global property of the entire embedding space, but is instead driven by a small number of idiosyncratic dimensions.

\subsubsection{Setup}
\citet{ethayarajh-2019-contextual} defines the anisotropy in layer $\ell$ of model $f$ as the expected cosine similarity of any pair of words in a corpus.
This can be approximated as $\textit{$\hat{A}$} (f_{\ell})$ from a sample $S$ of $n$ random token pairs from a corpus $\mathcal{O}.$ $S = \{\{x_1,y_1\},...,\{x_n,y_n\}\} \sim$ $\mathcal{O}$:
\begin{equation}
\begin{split}
     \textit{$\hat{A}$} (f_{\ell}) &= \frac{1}{n}\ \cdot\!\!\!\!\!\! \sum_{\{x_\alpha,y_\alpha\} \in S} \cos( f_{\ell}(x_\alpha), f_{\ell}(y_\alpha) )
\end{split}
\end{equation}

The cosine similarity, between two vectors $u$ and $v$ of dimensionality $d$ is defined as
\begin{equation}
    \cos(u, v) =
    \frac{u \cdot v}
    {{\lVert u \rVert}{\lVert v \rVert}} =
    \sum_{i=1}^d \frac{ u_i v_i}{{\lVert u \rVert}{\lVert v \rVert}}
\end{equation}

Expressing cosine similarity as a summation over $d$ dimensions, we can define a function $CC_i( u, v)$ which gives contribution of dimension $i$ to the total cosine similarity of $u$ and $v$ as:
\begin{equation}
\begin{split}
    CC_i(u, v) =
    \frac{ u_i v_i}{{\lVert u \rVert}{\lVert v \rVert}}
\end{split}
\end{equation}

From this, we define ${CC}(f^i_{\ell})$, the contribution of dimension $i$ to $\hat{A} (f_{\ell})$ as:

\begin{equation}
\begin{split}
    {CC}(f^i_{\ell}) = 
     \frac{1}{n} \ \cdot\!\!\!\!\!\!\sum_{\{x_\alpha,y_\alpha\} \in S} CC_i( f_{\ell}(x_\alpha), f_{\ell}(y_\alpha) )
\end{split}
\end{equation}
Note that $\sum^d_i{CC}(f^i_{\ell}) = \hat{A}(f_{\ell})$. 
From the mean cosine contribution by dimension, we can determine how much each dimension contributes to the total anisotropy. If ${CC}(f^1_{\ell}) \approx {CC}(f^2_{\ell})\approx...\approx {CC}(f^d_{\ell})$ then we conclude that anisotropy is a global property of the embedding space; no one dimension drives the expected cosine similarity of any two embeddings. By contrast, if ${CC}(f^i_{\ell}) >> \sum^d_{j \neq i}{CC}(f^j_{\ell})$ then we conclude that dimension $i$ dominates the cosine similarity computation.

\subsubsection{Experiment}
We compute the average cosine similarity contribution, ${CC}(f^i_{\ell})$, for each dimension in all layers of BERT, RoBERTa, GPT-2, and XLNet.\footnote{All models from \url{https://github.com/huggingface/transformers}} We then normalize by the total expected cosine similarity $\hat{A} (f_{\ell})$ to get the proportion of the total expected cosine similarity contributed by each dimension. All models are of dimensionality $d=768$ and have 12 layers, plus one static embedding layer. We also include two 300 dimensional non-contextual models, Word2Vec\footnote{\url{https://zenodo.org/record/4421380}} and GloVe,\footnote{\url{https://nlp.stanford.edu/projects/glove/} (Wikipedia+Gigaword 5, 300d)} for comparison. Our corpus {$\mathcal{O}$} is an 85k token sample of random articles from English Wikipedia. All input sequences consisted of 128 tokens. From the resulting representations we take a random sample $S$ of 500k token pairs. For each model, we report the three dimensions with the largest cosine contributions in the two most anisotropic layers, as well as the overall anisotropy $\hat{A}(f_{\ell})$.

\begin{table}
\small
\centering
\begin{tabular}{cccccc}
\hline
\textbf{Model} & \textbf{Layer} & \textbf{1} & \textbf{2} & \textbf{3} & \textbf{$\textit{$\hat{A}$} (f_{\ell})$}\\
\hline
GPT-2  & 11 & 0.275 & 0.269 & 0.265 & 0.640 \\
       & 12 & 0.763 & 0.131 & 0.078 & 0.885 \\
         \hline
BERT    & 10 & 0.817 & 0.004 & 0.003 & 0.396 \\
        & 11 & 0.884 & 0.003 & 0.002 & 0.506 \\
         \hline
RoBERTa & 7 & 0.726 & 0.193 & 0.032 & 0.705 \\
        & 12 & 0.663 & 0.262 & 0.020 & 0.745 \\
         \hline
XLNet  & 10 & 0.990 & 0.000 & 0.000 & 0.887 \\
       & 11 & 0.996 & 0.001 & 0.000 & 0.981 \\
         \hline
Word2Vec   & & 0.031 & 0.023 & 0.023 & 0.130 \\
GloVe      & & 0.105 & 0.096 & 0.095 & 0.104 \\
\end{tabular}
\caption{Proportion of total expected cosine similarity, ${CC}(f^i_{\ell})/\textit{$\hat{A}$} (f_{\ell})$, contributed by each of the top 3 dimensions in the two most anisotropic layers of each model, along with the anisotropy estimate {$\textit{$\hat{A}$} (f_{\ell})$} for the given layer. Results for all layers can be found in Table \ref{tab:cc_full} of the appendix.}
\label{tab:cc}
\end{table}

\subsubsection{Results and Discussion}
Results are summarized in Table \ref{tab:cc}. The static models Word2Vec and GloVe are relatively isotropic and are not dominated by any single dimension. Across all transformer models tested, \textbf{a small subset of rogue dimensions dominate the cosine similarity computation}, especially in the more anisotropic final layers. Perhaps the most striking case is layers 10 and 11 of XLNet, where a single dimension contributes more than 99\% of the expected cosine similarity between randomly sampled tokens.

The dimensions which drive anisotropy are centered far from the origin relative to other dimensions. For example, the top contributing dimension in the final layer of XLNet ($i=667$) has a mean activation of $\mathbb{E}[x^{667}_{12}] = 180.0$, while the expected activation of all other dimensions is $\mathbb{E}[x^{i\neq667}_{12}] = -0.084$ with standard deviation $\sigma[x^{i\neq667}_{12}] = 0.77$.

One implication of anisotropy is that the embeddings occupy a narrow cone in the embedding space, as the angle between any two word embeddings is very small. However, if anisotropy is driven by a single dimension (or a small subset of dimensions), we can conclude that the cone lies along a single axis or within a low dimensional subspace, rather than being a global property across all dimensions.%
\footnote{Our analysis complements that of \citet{cai2021isotropy} which used Principle Component Analysis to identify isolated isotropic clusters as well as embedding cones in a space reduced to three dimensions.} We conclude from this analysis that the anisotropy of the embedding space is an artifact of cosine similarity's high sensitivity to a small set of outlier dimensions and is not a global property of the space.%
\footnote{We additionally replicated \citet{ethayarajh-2019-contextual} before and after removing rogue dimensions in Appendix \ref{sec:appendixA}. We show that their analyses are extremely sensitive to rogue dimensions.}

\subsection{Informativity of Similarity Measures}
\label{sec:informativitycos}
In the previous section, we found that anisotropy is driven by a small subset of dimensions. In this section, we investigate whether standard similarity measures are still informed by the entire embedding space, or if \emph{variability} in the measure is also driven by a small subset of dimensions.

For example, it could be the case that some dimension $i$ has a large, but roughly constant activation across all tokens, meaning $\mathbb{E}[{CC}(f^i_{\ell})]$ will be large, but $Var[{CC}(f^i_{\ell})]$ will be near zero. In this case, we would be adding a large constant to cosine similarity, making $\textit{Anisotropy} (f_{\ell})$ large but not changing $Var[{cos}(f_{\ell}(x), f_{\ell}(y)]$. In this case, the average cosine similarity would be driven toward 1.0 by dimension $i$, but any changes in cosine similarity would be driven by the rest of the embedding space, not dimension $i$, meaning cosine similarity would provide information about the entire representation space, rather than a single dimension.
Conversely, dimension $i$ may have mean activation near zero, but extremely large variance across tokens. In this case, dimension $i$ would not appear to make the space anisotropic, but would still drive variability in cosine similarity. 
Ultimately, we're not interested in where the representation space is centered, but whether changes in a similarity measure reflect changes in the entire embedding space.

In this section we uncover which dimensions drive the \textit{variability} of cosine similarity.\footnote{We conduct the same analysis using Euclidean distance in Appendix \ref{sec:appdxL2} and reach similar conclusions as with cosine similarity.} Paralleling our findings in Section \ref{sec:2} we find that the token pairs which are similar/dissimilar to one another completely change when we remove just 1-5 dominant dimensions from the embedding space.

\subsubsection{Setup}
Let $f_{\ell}(x) : X \xrightarrow{}\mathbb{R}^d$, be the function which maps a token $x$ to its representation in layer $l$ of model $f$. Let $f'_{\ell}(x) : X \xrightarrow{}\mathbb{R}^{d-k}$ be the function which maps token $x$ to its representation with top k dimensions (measured by contribution to cosine similarity) removed. Let $C(S) = \underset{x,y \in S}{cos}(f_{\ell}(x), f_{\ell}(y))$ and $C'(S) = \underset{x,y \in S}{cos}(f'_{\ell}(x), f'_{\ell}(y))$. In this analysis, we compute:

\begin{equation}
r = Corr[C(S), C'(S)]
\end{equation}

This is the Pearson correlation between the cosine similarities in the entire embedding space and those similarities when k dimensions are removed. In our analysis we report $r^2$ which corresponds to the proportion of variance in $C(S)$ explained by $C'(S)$. For example, if we were to set k=1, and the observed $r^2$ is large, then cosine similarity in the full embedding space is still well explained by the remaining $d-1$ dimensions. By contrast, if $r^2$ is small, then the variance of cosine similarity in the embedding space can not be well explained by the bottom $d-1$ dimensions, and thus a single dimension drives variability in cosine similarity.

\subsubsection{Experiment}
For this experiment, we compute $r^2 = Corr[C(x,y), C'(x,y)]^2$ for all layers of all models, using the same set of token representations as in Section \ref{sec:2}. We remove the top $k=\{1,3,5\}$ dimensions, where dimensions are ranked by ${CC}(f^i_{\ell})$, the cosine similarity contribution of dimension $i$ in layer $l$. We report results for the first layer and the final two layers. Results for all layers can be found in Table \ref{table:allvar} of the Appendix.

\begin{table}
\small
\centering
\begin{tabular}{cccccc}
\hline
\textbf{Model} & \textbf{Layer} & \textbf{k=1} & \textbf{k=3} & \textbf{k=5}\\
\hline
GPT-2   & 0 & 0.999 &  0.996 &  0.996 \\
        & 11 & 0.967 &  0.352 &  0.352 \\
        & 12 & 0.819 &  0.232 &  0.232 \\
         \hline
BERT    & 0 & 0.999 &  0.997 &  0.997 \\
        & 11 & 0.046 &  0.048 &  0.048 \\
        & 12 & 0.213 &  0.214 &  0.214 \\
         \hline
RoBERTa & 0 & 0.810 &  0.770 &  0.770 \\
        & 11 & 0.591 &  0.319 &  0.319 \\
        & 12 & 0.566 &  0.301 &  0.301 \\
         \hline
XLNet   & 0 & 0.999 &  0.996 &  0.996 \\
        & 11 & 0.124 &  0.150 &  0.150 \\
        & 12 & 0.028 &  0.024 &  0.024 \\
         \hline
Word2vec   & & 0.998 &  0.993 &  0.988 \\
GloVe        & & 0.987 &  0.954 & 0.930 \\
\end{tabular}
\caption{Proportion of variance in cosine similarity \textbf{$r^2$} explained by cosine similarity when the top $k$ dimensions, measured by ${CC}(f^i_{\ell})$, are removed. Layer 0 is the static embedding layer. Results for all layers can be found in Table \ref{table:allvar} of the Appendix.}
\label{tab:cosinf}
\end{table}

\subsubsection{Results}
Results are summarized in Table \ref{tab:cosinf}. We find that in the static embedding models and the earlier layers of each contextual model, no single dimension or subset of dimensions drives the variability in cosine similarity. By contrast, in later layers, the variability of cosine similarity is driven by just 1-5 dimensions. In the extreme cases of XLNet-12 and BERT-11, when we remove just a single dimension from the embedding space, almost none of the variance in cosine similarity can be explained by cosine similarity in the $d-1$ dimensional subspace. ($r^2 =$ 0.028 and 0.046 respectively) This means that the token pairs which are similar to one another in the full embedding space are drastically different from the pairs which are similar when just a handful of dimensions are removed.

While similarity measures should reflect properties of the entire embedding space, we have shown that this is not the case with cosine similarity in contextualized embedding spaces. Not only do a small subset of dimensions in later layers drive the cosine similarity of randomly sampled words toward 1.0, but this subset also drives the variability of the measure. \textbf{This result effectively renders cosine similarity a measure over 1-5 rogue dimensions rather than the entire embedding space.}

\begin{figure*}[h!]
    \includegraphics[width=\linewidth]{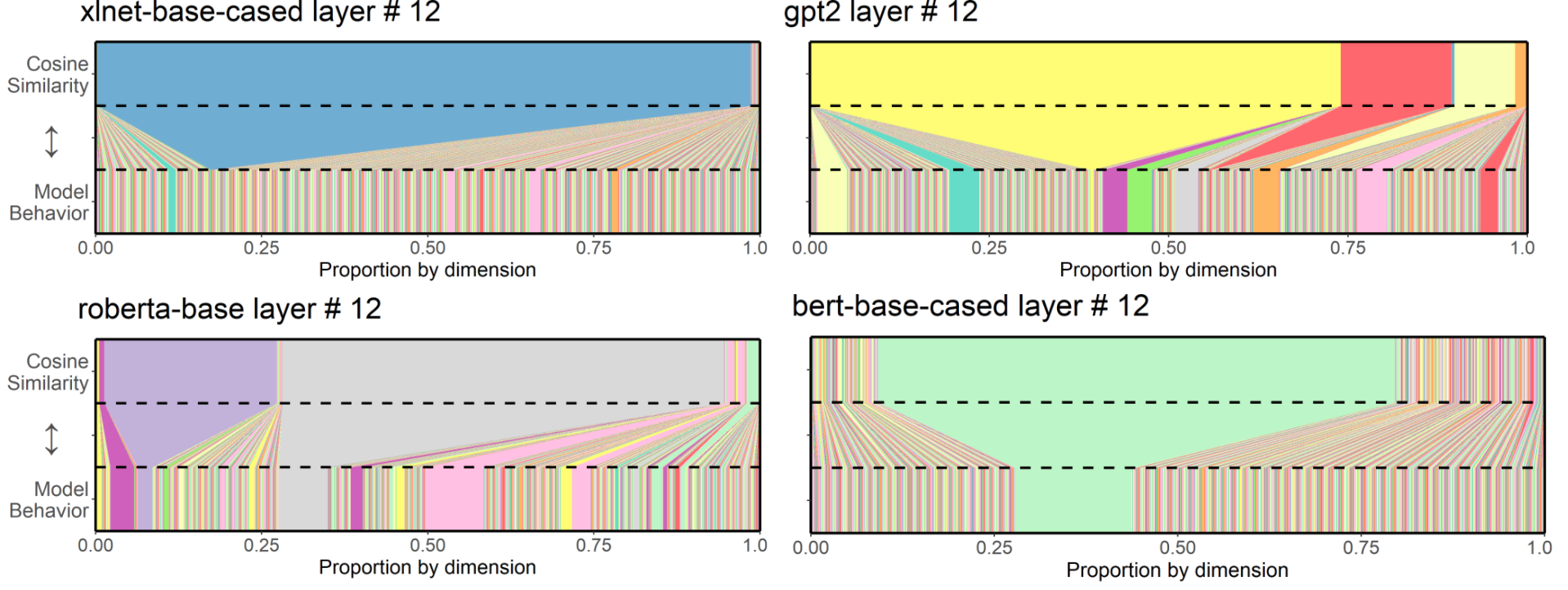}
    \caption{Relative contribution of each dimension to cosine similarity, ${CC}(f^i_{\ell})$, (top) paired with its relative influence on model behavior, $I(i,\ell, f)$ (bottom). The top and bottom portions of the plots each have 768 bars, one for each dimension in layer 12. The width of the bars corresponds to their relative contribution to each metric. For example, three dimensions (yellow, red, light yellow) dominate cosine similarity in GPT-2, but when we trace those dimensions to the bottom half of the plot, they appear to vanish, meaning their relative influence on model behavior is negligible. While this mismatch is less pronounced for BERT, it is particularly extreme in XLNet, where a single dimension dominates cosine similarity, but is effectively meaningless to the pretraining objective.}
    \label{fig:behavior}
\end{figure*}

\section{Rogue Dimensions and Model Behavior}
\label{sec:behavior}
In this section, we address the question of whether the dimensions which dominate cosine similarity likewise dominate model behavior. Specifically, if similarity measures are dominated by only a few dimensions, as shown in the previous sections, then those dimensions should be the only ones the model actually uses, otherwise, the measures only reflect a small subset of what the model is doing. We find that dimensions which dominate cosine similarity do not likewise dominate model behavior.

\subsection{Behavioral Influence of Individual Dimensions}
We measure the influence of individual dimensions on model behavior through an ablation study in the style of \citet{s.2018on}.%
\footnote{There are several possible ways to assess the importance of individual neurons on prediction. One popular technique is Layerwise Relevance Propagation \cite{bach-lrp} which has recently been used in Transformer-based models \cite{voita2020analyzing}. We use feature ablation due to its ease of implementation and generalizability across architectures.} The idea of neuron ablation studies is to examine how the performance of a network changes when a neuron is clamped to a fixed value, typically zero. In our study, we measure how much the language modeling distribution changes when dimension $i$ of layer $\ell$ is fixed to zero.

\subsection{Setup}
Let $P_f(s)$ be the original language modeling distribution of model $f$ for some input $s$ sampled from corpus $\mathcal{O}$. We measure how the distribution changes after ablation using KL divergence between the ablated model distribution and the unaltered reference distribution.%
\footnote{We zero out dimensions by setting the appropriate layer normalization parameters $\gamma$ and $\beta$ to zero.} We use KL divergence, rather than typical measures of importance in feature ablation such as accuracy or perplexity because we are interested in how much the prediction distributions change rather than performance on some task. Our measure of the importance of dimension $i$ in layer $\ell$ of model $f$ is the mean KL divergence between the two distributions across our corpus, where $S$ is a set of $n$ inputs to the model.

\begin{equation}
\begin{split}
I(i,\ell, f) &=
\frac{1}{n} \sum_{s \in S}^n D_{KL}[P_f(s)  \rVert  P_{f}(s|f^i_{\ell}(s) = 0)]
\end{split}
\end{equation}

\subsection{Experiment}
To measure the importance of each dimension to model behavior, we compute $I(i,\ell, f)$ for the last 4 layers of each model over 10k distributions. Since the autoregressive models (GPT-2, XLNet) give a language modeling distribution over all tokens in the input, we use a corpus of 10k tokens from English Wikipedia. In the auto-encoder models (BERT, RoBERTa), we mask 15\% of tokens and use a corpus of 150k tokens, for a total of 10k language modeling distributions. We plot the relative behavioral influence of each dimension against its contribution to cosine similarity, measured by ${CC}(f^i_{\ell})$, (each is normalized to sum to 1).

\subsection{Results}
Figure \ref{fig:behavior} displays the results for the final layer of each model.%
\footnote{The plots for layers 9-11 can be found in Figure \ref{fig:mismatch_9-11} in the supplementary materials.}
In all models, we see that \textbf{the dimensions which dominate cosine similarity do not likewise dominate model behavior.} The mismatch is less drastic in BERT's final layer, but is quite severe in final XLNet and GPT-2, where removing the dimensions which dominate cosine similarity does not lead to substantial changes in the language modeling distribution. 

While ablating rogue dimensions often alters the language modeling distribution more than ablating non-rogue dimensions, we emphasize that there is not a one-to-one correspondence between a dimension's influence on cosine similarity and its influence on language modeling behavior. In the case of XLNet and GPT-2, removing dimensions which dominate cosine similarity leads to only vanishingly small changes to the behavior of the model.

\subsection{Behavioral Correlates of Rogue Dimensions}

We now turn to the related question of whether rogue dimensions actually capture linguistically meaningful information. Because rogue dimensions dominate representational similarity measures, these measures will be heavily biased toward whatever information these dimensions capture. To explore their behavioral correlates, we plotted the distribution of the values for rogue dimensions.

\begin{figure*}[h!]
    \includegraphics[width=\linewidth]{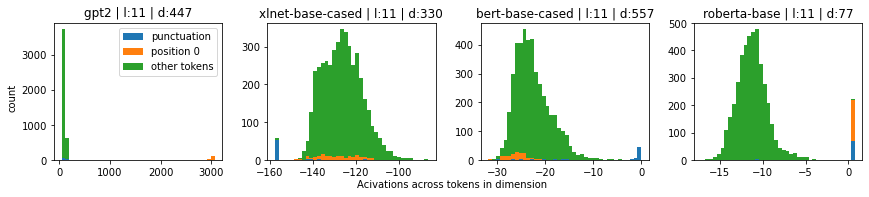}
    \caption{Distribution of values in the dimension with the highest variance in layer 11 of each model across a sample of 10k tokens from English Wikipedia. Each color corresponds to a specific type/position. The orange distribution is tokens which occur in position zero, the blue distribution is instances of the "." token, and green is instances of all other tokens. Results for all layers can be found in Figures \ref{fig:rogue_dim_0} and \ref{fig:rogue_dim_1} of the appendix.}
    \label{fig:behavior_corr}
\end{figure*}

We show in Figure \ref{fig:behavior_corr} that rogue dimensions often have highly type/position specific activation patterns. Rogue dimensions in all models are particularly sensitive to instances of the "." token and/or position 0 of the input. For example, in laters 2-11 of GPT-2 and RoBERTa, the mean cosine similarity of any two tokens in position 0 is greater than .99, while the mean similarity of tokens not in position 0 is .623 and .564 respectively. 

While the transformer language models we have tested have all been shown to capture a rich range of linguistic phenomena, this linguistic knowledge may be obscured by rogue dimensions. The following section empirically evaluates this hypothesis.

\section{Postprocessing and Representational Quality}
\label{sec:repquality}
While we have shown that the representational geometry of contextualized embeddings makes cosine similarity uninformative, there are several simple postprocessing methods which can correct for this. In this section we outline three such methods: standardization, all-but-the-top \cite{mu2018allbutthetop}, and ranking (via Spearman correlation). We evaluate representational quality of the postprocessed embeddings on several word similarity/relatedness datasets and show that the underlying representational quality is obscured by the rogue dimensions. When we correct for rogue dimensions, correlation with human similarity judgments improves across the board. We also find that representational quality is preserved across all layers, rather than giving way to degraded/task specific representations as argued in previous work.

\subsection{Postprocessing}
\textbf{Standardization}: We have observed that a small subset of dimensions with means far from zero and high variance completely dominate cosine similarity. A straightforward way to adjust for this is to subtract the mean vector and divide each dimension by its standard deviation, such that each dimension has $\mu_i = 0$ and $\sigma_i = 1$. Concretely, given some corpus of length $|\mathcal{O}|$ containing word representations $x \in \mathbb{R}^d$, we compute the mean vector $\mu \in \mathbb{R}^d$
\begin{equation}
\mu = \frac{1}{|\mathcal{O}|} \cdot \sum_{x \in \mathcal{O}} x
\end{equation}

as well as the standard deviation in each dimension $\sigma \in \mathbb{R}^d$
\begin{equation}
\sigma = \sqrt{\frac{1}{|\mathcal{O}|} \cdot \sum_{x \in \mathcal{O}} (x - \mu)^2}
\end{equation}

Our new standardized representation for each word vector ($z$) becomes the z-score in each dimension.
\begin{equation}
z = \frac{x - \mu}{\sigma}
\end{equation}

\textbf{All-but-the-top}: Following from similar observations (a nonzero common mean vector and a small number of dominant directions) in static embedding models, \citet{mu2018allbutthetop} proposed subtracting the common mean vector and eliminating the top few principle components (they suggested the top $\frac{d}{100}$), which should capture the variance of the rogue dimensions in the model%
\footnote{See \citet{cai2021isotropy} for further discussion of the top principle components of contextual language models.} and make the space more isotropic.

\textbf{Spearman's $\rho$}: \citet{zhelezniak-etal-2019-correlation} treat word embeddings as $d$ observations from an $|\mathcal{O}|$-variate distribution, and use Pearson correlation as a measure of similarity. They propose the use of non-parametric rank correlation coefficients, such as Spearman's $\rho$ when embeddings depart from normality. Spearman correlation is just Pearson correlation but between the ranks of embeddings, rather than their values. Thus Spearman correlation can also be thought of as a postprocessing technique, where instead of standardizing the space or removing the top components, we simply transform embeddings as
$x' = \textit{rank(x)}$. Spearman's $\rho$ is robust to outliers and thus will not be dominated by the rogue dimensions of contextual language models. Unlike standardization and all-but-the-top, Spearman correlation requires no computations over the entire corpus. While rank-based similarity measures will not be dominated by rogue dimensions, rogue dimensions will tend to occupy the top or bottom ranks.

\subsection{Representational Quality}
While we have shown that cosine similarity is dominated by a small subset of dimensions, a remaining question is whether adjusting for these dimensions makes similarity measures more informative. In particular, we evaluate whether the cosine similarities between word pairs align more closely with human similarity judgments after post-processing. We evaluate this using 4 word similarity/relatedness judgment datasets: RG65 \cite{10.1145/365628.365657}, WS353 \cite{agirre-etal-2009-study}, {\sc{SimLex999}} \cite{hill-etal-2015-simlex} and {\sc{SimVerb3500}} \cite{gerz-etal-2016-simverb}. Examples in these datasets consist of a pair of words and a corresponding similarity rating averaged over several human annotators. Because the similarity judgments were designed to evaluate static embeddings, we use the context-aggregation strategy of \citet{bommasani-etal-2020-interpreting} to produce static representations.%
\footnote{We aggregate over between 200-500 single-sentence contexts of each word type using sentences from English Wikipedia. Words with an insufficient number of contexts were omitted, leaving a total of 1,894 unique words and 4,577 unique pairs. We use mean pooling over subwords to get a single representation for a word.}

For each model, we report the Spearman correlation between the model similarities and human-similarity judgments, averaged across all 4 datasets.%
\footnote{Full results from each dataset can be seen in Figures \ref{fig:rep_qualitygpt}, \ref{fig:rep_qualitybert}, \ref{fig:rep_qualityroberta}, \ref{fig:rep_qualityxlnet} of the Appendix.}
We report the correlation for cosine similarities of the original embeddings, as well as for postprocessed embeddings using four strategies: standardization, all-but-the-top (removing the top 7 components), only subtracting the mean (the step common to both strategies) and Spearman correlation.

\begin{figure*}[h!]
    \includegraphics[width=\linewidth]{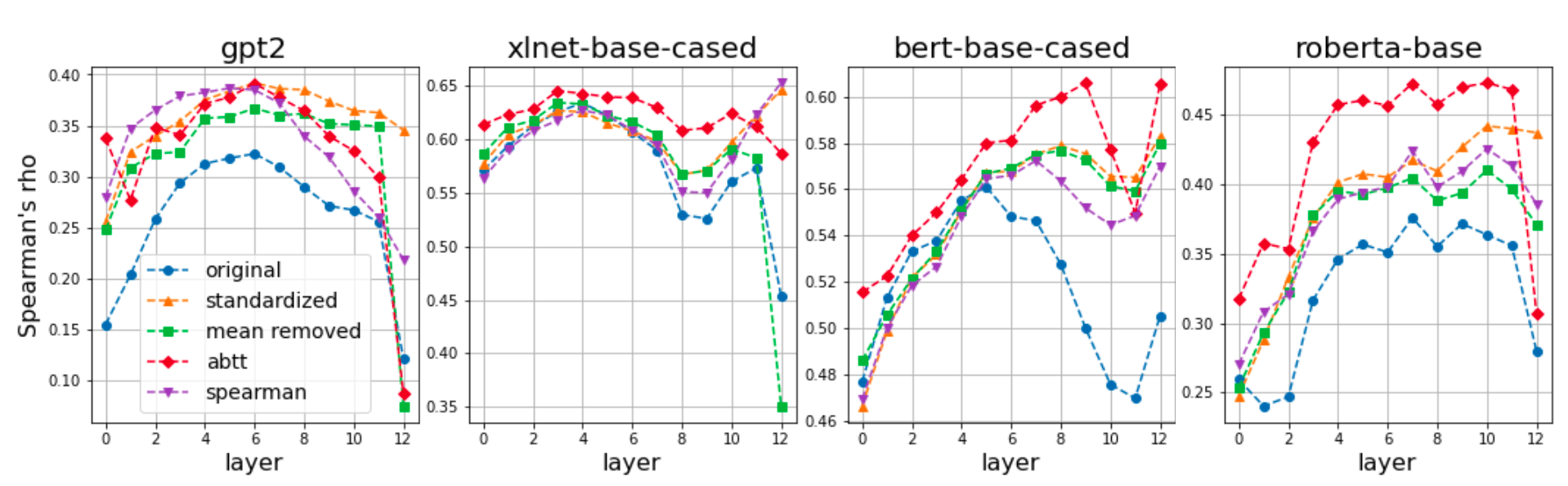}
    \caption{Average correlation (Spearman's $\rho$) with human judgments in the four word similarity datasets, with and without postprocessing.}
    \label{fig:quality}
\end{figure*}

\subsection{Results}
Results are summarized in Figure \ref{fig:quality}. Our key findings are:

\textbf{Postprocessing aligns the embedding space more closely to human similarity judgments} across almost all layers of all models. We found that standardization was the most successful postprocessing method, showing consistent improvement over the original embeddings in all but the early layers of BERT. 

All-but-the-top was generally effective, though the resulting final layer of RoBERTa and GPT-2 exhibited poor correlation with human judgements, similar to the original embeddings. In pilot analyses, we found that all-but-the-top is highly dependent on the number of components removed, a hyperparameter, $D$, which \citet{mu2018allbutthetop} suggest should be $\frac{d}{100}$. Just removing the first principle component in RoBERTa yielded a stronger correlation, but all-but-the top did not significantly improve correlation with human judgements in the final layer of GPT-2 for any choice of $D$. 

Simply subtracting the mean vector also yielded substantial gains in most models with the exception of the final layers of GPT-2 and XLNet. The rogue dimensions in the last layer of these two models have exceptionally high variance. While subtracting the mean made the space more isotropic as measured by cosine similarity, it did not reduce the variance of each dimension. We found, particularly in the final layer of GPT-2 and XLNet that 1-3 dimensions drive the variability of cosine similarity, and this was still the case when the mean vector was subtracted.

Converting embeddings into ranks (Spearman correlation) also resulted in significantly stronger correlations with human judgments in all layers of all models, though the correlation was often weaker than standardization or all-but-the-top.

\textbf{Representational quality is preserved across all layers.}
Previous work has suggested that the final layers of transformer language models are highly task-specific. \citet{liu-etal-2019-linguistic} showed that the middle layers of BERT outperform the final layers on language understanding tasks. Using a cosine-similarity based text-generation evaluation metric, \citet{Zhang*2020BERTScore:} showed a sharp drop in correlation to human judgements of machine translation quality in final layers of various transformer language models. Similarly, \citet{davis-van-schijndel-2020-discourse} used Representational Similarity Analysis (RSA) with Pearson correlation%
\footnote{\citet{zhelezniak-etal-2019-correlation} showed Pearson correlation to be effectively equivalent to cosine similarity.} and found that intermediate layers of GPT-2 and TransformerXL encode human-like implicit causality biases which are subsequently obscured in final layers. 

Our findings suggest that linguistic representational quality (in this case lexical semantics) is actually preserved in the final layers but is obscured by a small handful of rogue dimensions. After simple postprocessing, later layers of the model correlate just as well, if not better than intermediate layers with human similarity judgments. This finding reaffirms the need to carefully consider the representational geometry of a model before drawing conclusions about layerwise representational quality, and the general linguistic knowledge these models encode.

\section{Discussion and Future Work}

Perhaps the most important direction for future work is designing and implementing language models which do not develop rogue dimensions in the first place. \citet{gao2018representation} introduce a cosine-regularization term during pretraining which improved the performance of transformer models on machine translation. Perhaps BERT or GPT models could similarly benefit from such regularization. 

A prerequisite for designing models without rogue dimensions is understanding how these dimensions arise over time. Contemporaneous work from \citet{bis-etal-2021-much} provides a useful characterization of how degenerate representations may be learned, which largely focuses on token frequency, while \citet{kovaleva-etal-2021-bert} provide a characterization of how outliers impact model performance, attributing much of the problem to scaling factors in layer normalization, and \citet{luo-etal-2021-positional} make observations about the contribution of positional embeddings. In the present work, we observe strong correlations with specific tokens and positions. Unifying these accounts is an important task for future work. With the recent release of the MultiBERT checkpoints \cite{sellam2021multiberts}, future work can uncover whether rogue dimensions are a coincidental property of some models, or whether they are a requisite for good performance. The MultiBERTs may also elucidate how these dimensions emerge during pretraining. While we empirically motivate a trivially simple transformation which corrects for rogue dimensions, we believe the most fruitful direction for future work is to build models whose representations require no post-hoc transformations. This would result in more interpretable embedding spaces and may additionally lead to models with better performance.

\section{Conclusion}
In this work, we showed that similarity measures in contextual language models are largely reflective of a small number of rogue dimensions, not the entire embedding space. Consequently, a few dimensions can drastically change the conclusions we draw about the linguistic phenomena a model actually captures. We showed that the previously observed anisotropy in contextual models is essentially an artifact of rogue dimensions and is not a global property of the entire embedding space. We also showed that variability in similarity is driven by just 1-5 dimensions of the embedding space. In many cases, removing just a single dimension completely changed which token pairs were similar to one another. However, we found that model behavior was not driven by these rogue dimensions, and that these dimensions seem to handle a small subset of a model's linguistic abilities, such as punctuation and positional information. In summary, standard similarity measures such as cosine similarity and Euclidean distance are not informative measures of how contextual language models represent and process language. We argue that measures of similarity in contextual language models must account for rogue dimensions using techniques such as standardization. These techniques should not just be viewed as avenues to improve downstream performance, but as prerequisites for any analysis involving representational similarity.

\section*{Acknowledgements}
We would like to thank Maria Antoniak, Valts Blukis, Forrest Davis, Liye Fu, Ge Gao, Tianze Shi, Ana Smith, Karen Zhou, members of the Cornell NLP Group and the Computational Psycholinguistics Discussions research group (C.Psyd) for their valuable feedback on earlier drafts of this work. We additionally thank Rishi Bommasani for productive, stimulating discussion. Finally, we thank the reviewers and area chairs for their detailed and insightful feedback. 

\bibliography{custom}
\bibliographystyle{acl_natbib}

\appendix

\section{Removing Dominant Dimensions and Representational Geometry}
\label{sec:appendixA}
\begin{figure*}[h!]
    \includegraphics[width=\linewidth]{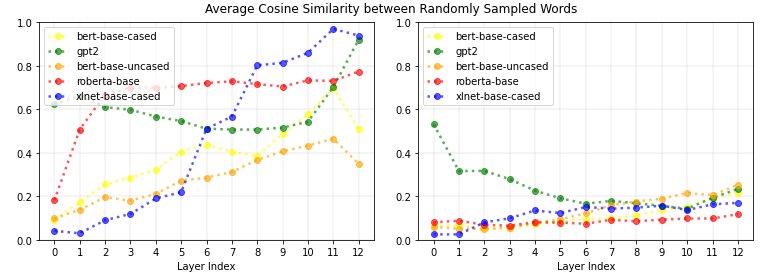}
    \caption{Anisotropy by layer of the full embedding space (left) and with the top 5 dimensions removed, as measured by $\mathbb{E}[CC_i]$ (right). In all models, anisotropy drastically decreases, and becomes more consistent across models and layers.}
    \label{fig:aniso}
\end{figure*}

\begin{figure*}[h!]
    \includegraphics[width=\linewidth]{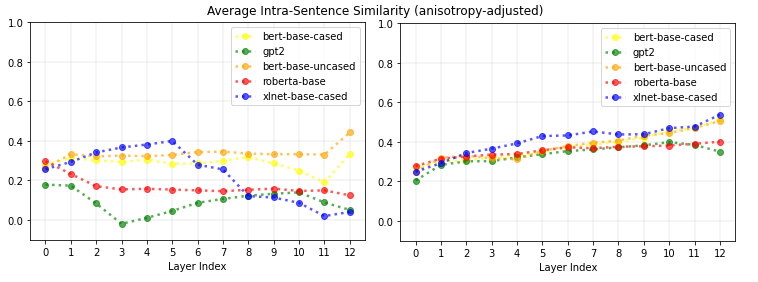}
    \caption{Intra-sentence similarity by layer of the full embedding space (left) and with the top 5 dimensions removed, as measured by $\mathbb{E}[CC_i]$ (right). Intra-sentence similarity is much more consistent and monotonically increasing when the top 5 dimensions are removed.}
    \label{fig:iss}
\end{figure*}

\begin{figure*}[h!]
    \includegraphics[width=\linewidth]{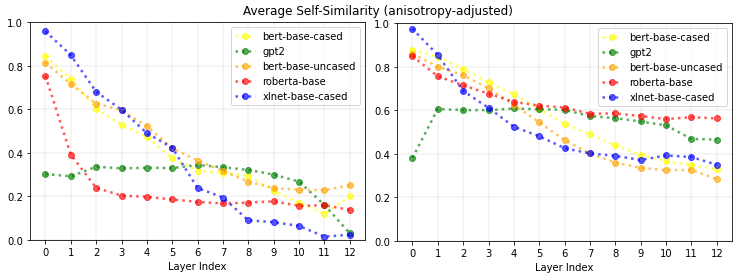}
    \caption{Average self-similarity (similarity of the same word type across contexts) by layer of the full embedding space (left) and with the top 5 dimensions removed, as measured by $\mathbb{E}[CC_i]$ (right). In the full embedding space, words of the same type in GPT-2 and XLNet appear no more similar to one another than randomly-sampled tokens. When we remove just 5 dimensions, words of the same type are indeed more similar to one another than the random baseline.}
    \label{fig:iss}
\end{figure*}

To facilitate a direct comparison with anisotropy estimates of \citet{ethayarajh-2019-contextual}, we replicate the experiments of Section 4 before and after removing the top $k$ dimensions with the largest $\mathbb{E}[CC_i]$. For these experiments we chose $k$=5 dimensions to remove. Results for anisotropy estimates are shown in Figure~\ref{fig:aniso}. Three key takeaways from this analysis are:

\textbf{All models tested had highly anisotropic representations}, including XLNet and RoBERTa which had not been evlauated in previous work. XLNet is even more anistropic than GPT-2 in its final two layers. RoBERTa's word representations are likewise highly anisotropic, though starting in earlier layers than in XLNet and BERT.

\textbf{After removing just 5 dimensions, embeddings become relatively isotropic}, with $\textit{$\hat{A}$} (f_{\ell})$ never larger than 0.25 in any layer of any model.

\textbf{Anisotropy becomes consistent across models and across layers}, suggesting that the deviant dimensions that drive anisotropy are idiosyncratic and model/layer specific; we show this to indeed be the case in Section \ref{sec:behavior}. By contrast, the geometry of the embedding space without rogue dimensions show similar properties across models/layers, suggesting that the similar qualities of the representational geometries of each model are obscured by these rogue dimensions. 

This can additionally be seen in our replication of the intra-sentence similarity and self-similarity from \citet{ethayarajh-2019-contextual}. While they find extreme cases in which words of the same type are no more similar to one another than randomly sampled words, we find a consistently high degree of self-similarity across all layers of all models after removing 5 dimensions. This suggests that information about word identity is preserved across all layers, rather than giving way to extremely contextualized representations in the final layer, this concurs with our findings in Section \ref{sec:repquality}. Together, these show that our conclusions about the geometry of contextual embedding spaces are heavily skewed by the sensitivity of cosine similarity to rogue dimensions present in each of these models.

\section{Informativity of Euclidean Distance}
\label{sec:appdxL2}
In this section, we conduct a similar analysis to Section \ref{sec:informativitycos} to see whether the variability in Euclidean distances between pairs of embeddings can be explained by Euclidean distance with the top $k$ dimensions are removed. Our methods for this analysis are identical to those of Section \ref{sec:informativitycos}, except our criterion for choosing $k$ is the variance in each dimension.  Results are shown in Table \ref{tab:l2inf}. In the extreme case of XLNet, none of the variability in Euclidean distances can be explained by Euclidean distances when a single dimension is removed. This means that Euclidean distance in this layer is effectively a measure of a single dimension.

\begin{table}
\small
\centering
\begin{tabular}{cccccc}
\hline
\textbf{Model} & \textbf{Layer} & \textbf{k=1} & \textbf{k=3} & \textbf{k=5}\\
\hline
GPT-2  & 0 & 0.999 &  0.996 &  0.996 \\
        & 1 & 0.983 &  0.975 &  0.975 \\
        & 2 & 0.999 &  0.783 &  0.783 \\
        & 3 & 0.992 &  0.257 &  0.257 \\
        & 4 & 0.993 &  0.200 &  0.200 \\
        & 5 & 0.993 &  0.159 &  0.159 \\
        & 6 & 0.993 &  0.090 &  0.090 \\
        & 7 & 0.992 &  0.037 &  0.037 \\
        & 8 & 0.990 &  0.007 &  0.007 \\
        & 9 & 0.990 &  0.002 &  0.002 \\
        & 10 & 0.986 &  0.022 &  0.022 \\
        & 11 & 0.971 &  0.974 &  0.974 \\
        & 12 & 0.909 &  0.333 &  0.333 \\
         \hline
BERT    & 0 & 0.997 &  0.997 &  0.997 \\
        & 1 & 0.994 &  0.993 &  0.993 \\
        & 2 & 0.993 &  0.992 &  0.992 \\
        & 3 & 0.994 &  0.993 &  0.993 \\
        & 4 & 0.988 &  0.987 &  0.987 \\
        & 5 & 0.992 &  0.991 &  0.991 \\
        & 6 & 0.988 &  0.987 &  0.987 \\
        & 7 & 0.982 &  0.981 &  0.981 \\
        & 8 & 0.969 &  0.968 &  0.968 \\
        & 9 & 0.925 &  0.924 &  0.924 \\
        & 10 & 0.762 &  0.761 &  0.761 \\
        & 11 & 0.434 &  0.433 &  0.433 \\
        & 12 & 0.990 &  0.989 &  0.989 \\
         \hline
RoBERTa & 0 & 0.810 &  0.770 &  0.770 \\
        & 1 & 0.509 &  0.264 &  0.264 \\
        & 2 & 0.584 &  0.141 &  0.141 \\
        & 3 & 0.607 &  0.152 &  0.152 \\
        & 4 & 0.657 &  0.200 &  0.200 \\
        & 5 & 0.623 &  0.225 &  0.225 \\
        & 6 & 0.641 &  0.242 &  0.242 \\
        & 7 & 0.614 &  0.241 &  0.241 \\
        & 8 & 0.578 &  0.235 &  0.235 \\
        & 9 & 0.591 &  0.270 &  0.270 \\
        & 10 & 0.575 &  0.281 &  0.281 \\
        & 11 & 0.591 &  0.319 &  0.319 \\
        & 12 & 0.566 &  0.301 &  0.301 \\
         \hline
XLNet   & 0 & 0.999 &  0.996 &  0.996 \\
        & 1 & 1.000 &  1.000 &  1.000 \\
        & 2 & 1.000 &  0.987 &  0.987 \\
        & 3 & 0.993 &  0.992 &  0.992 \\
        & 4 & 0.983 &  0.978 &  0.978 \\
        & 5 & 0.903 &  0.896 &  0.896 \\
        & 6 & 0.481 &  0.470 &  0.470 \\
        & 7 & 0.432 &  0.426 &  0.426 \\
        & 8 & 0.235 &  0.236 &  0.236 \\
        & 9 & 0.321 &  0.323 &  0.323 \\
        & 10 & 0.308 &  0.307 &  0.307 \\
        & 11 & 0.124 &  0.150 &  0.150 \\
        & 12 & 0.028 &  0.024 &  0.024 \\
\end{tabular}
\caption{Proportion of variance in Euclidean distance \textbf{$r^2$} explained by Euclidean distance when the top $k$ dimensions (measured by the variance in each dimension) are removed.}
\label{tab:l2inf}
\end{table}

\begin{table}
\small
\centering
\begin{tabular}{cccccc}
\hline
\textbf{Model} & \textbf{Layer} & \textbf{1} & \textbf{2} & \textbf{3} & \textbf{$\textit{$\hat{A}$} (f_{\ell})$}\\
\hline
GPT-2   & 0 & 0.054 & 0.051 & 0.051 & 0.484 \\
        & 1 & 0.324 & 0.163 & 0.150 & 0.626 \\
        & 2 & 0.319 & 0.205 & 0.149 & 0.612 \\
        & 3 & 0.294 & 0.264 & 0.145 & 0.589 \\
        & 4 & 0.297 & 0.275 & 0.151 & 0.549 \\
        & 5 & 0.324 & 0.258 & 0.150 & 0.517 \\
        & 6 & 0.351 & 0.237 & 0.148 & 0.485 \\
        & 7 & 0.374 & 0.205 & 0.144 & 0.466 \\
        & 8 & 0.376 & 0.156 & 0.141 & 0.461 \\
        & 9 & 0.364 & 0.190 & 0.157 & 0.466 \\
        & 10 & 0.326 & 0.257 & 0.207 & 0.498 \\
        & 11 & 0.275 & 0.269 & 0.265 & 0.640 \\
        & 12 & 0.763 & 0.131 & 0.078 & 0.885 \\
         \hline
BERT    & 0 & 0.159 & 0.076 & 0.035 & 0.066 \\
        & 1 & 0.541 & 0.049 & 0.024 & 0.154 \\
        & 2 & 0.790 & 0.006 & 0.005 & 0.224 \\
        & 3 & 0.792 & 0.006 & 0.004 & 0.234 \\
        & 4 & 0.781 & 0.007 & 0.005 & 0.283 \\
        & 5 & 0.809 & 0.007 & 0.005 & 0.360 \\
        & 6 & 0.792 & 0.005 & 0.004 & 0.382 \\
        & 7 & 0.716 & 0.006 & 0.005 & 0.342 \\
        & 8 & 0.668 & 0.006 & 0.006 & 0.326 \\
        & 9 & 0.743 & 0.004 & 0.004 & 0.380 \\
        & 10 & 0.817 & 0.004 & 0.003 & 0.396 \\
        & 11 & 0.884 & 0.003 & 0.002 & 0.506 \\
        & 12 & 0.686 & 0.005 & 0.005 & 0.370 \\
         \hline
RoBERTa & 0 & 0.726 & 0.040 & 0.021 & 0.143 \\
        & 1 & 0.850 & 0.081 & 0.009 & 0.442 \\
        & 2 & 0.862 & 0.093 & 0.013 & 0.627 \\
        & 3 & 0.841 & 0.113 & 0.017 & 0.659 \\
        & 4 & 0.796 & 0.146 & 0.023 & 0.666 \\
        & 5 & 0.775 & 0.160 & 0.025 & 0.672 \\
        & 6 & 0.745 & 0.180 & 0.030 & 0.679 \\
        & 7 & 0.726 & 0.193 & 0.032 & 0.705 \\
        & 8 & 0.674 & 0.229 & 0.038 & 0.690 \\
        & 9 & 0.648 & 0.254 & 0.040 & 0.675 \\
        & 10 & 0.698 & 0.223 & 0.032 & 0.689 \\
        & 11 & 0.666 & 0.252 & 0.031 & 0.696 \\
        & 12 & 0.663 & 0.262 & 0.020 & 0.745 \\
         \hline
XLNet   & 0 & 0.300 & 0.043 & 0.028 & 0.037 \\
        & 1 & 0.085 & 0.059 & 0.036 & 0.022 \\
        & 2 & 0.042 & 0.031 & 0.016 & 0.050 \\
        & 3 & 0.157 & 0.013 & 0.011 & 0.051 \\
        & 4 & 0.413 & 0.017 & 0.009 & 0.169 \\
        & 5 & 0.700 & 0.005 & 0.004 & 0.177 \\
        & 6 & 0.908 & 0.003 & 0.002 & 0.514 \\
        & 7 & 0.942 & 0.001 & 0.001 & 0.563 \\
        & 8 & 0.982 & 0.000 & 0.000 & 0.826 \\
        & 9 & 0.984 & 0.000 & 0.000 & 0.833 \\
        & 10 & 0.990 & 0.000 & 0.000 & 0.887 \\
        & 11 & 0.996 & 0.001 & 0.000 & 0.981 \\
        & 12 & 0.973 & 0.003 & 0.002 & 0.884 \\
\end{tabular}
\caption{Proportion of total expected cosine similarity, ${CC}(f^i_{\ell})/\textit{$\hat{A}$} (f_{\ell})$, contributed by each of the top 3 dimensions for all layers of each model, along with the anisotropy estimate {$\textit{$\hat{A}$} (f_{\ell})$} for the given layer.}
\label{tab:cc_full}
\end{table}

\begin{table}
\small
\centering
\begin{tabular}{cccccc}
\hline
\textbf{Model} & \textbf{Layer} & \textbf{k=1} & \textbf{k=3} & \textbf{k=5}\\
\hline
GPT-2   & 0 & 0.999 &  0.996 &  0.996 \\
        & 1 & 0.985 &  0.888 &  0.888 \\
        & 2 & 0.990 &  0.899 &  0.899 \\
        & 3 & 0.991 &  0.849 &  0.849 \\
        & 4 & 0.910 &  0.775 &  0.775 \\
        & 5 & 0.872 &  0.719 &  0.719 \\
        & 6 & 0.853 &  0.684 &  0.684 \\
        & 7 & 0.862 &  0.713 &  0.713 \\
        & 8 & 0.894 &  0.797 &  0.797 \\
        & 9 & 0.921 &  0.490 &  0.490 \\
        & 10 & 0.947 &  0.428 &  0.428 \\
        & 11 & 0.967 &  0.352 &  0.352 \\
        & 12 & 0.819 &  0.232 &  0.232 \\
         \hline
BERT    & 0 & 0.999 &  0.997 &  0.997 \\
        & 1 & 0.894 &  0.848 &  0.848 \\
        & 2 & 0.580 &  0.568 &  0.568 \\
        & 3 & 0.514 &  0.504 &  0.504 \\
        & 4 & 0.459 &  0.449 &  0.449 \\
        & 5 & 0.383 &  0.374 &  0.374 \\
        & 6 & 0.343 &  0.338 &  0.338 \\
        & 7 & 0.391 &  0.394 &  0.394 \\
        & 8 & 0.400 &  0.398 &  0.398 \\
        & 9 & 0.219 &  0.220 &  0.220 \\
        & 10 & 0.119 &  0.123 &  0.123 \\
        & 11 & 0.046 &  0.048 &  0.048 \\
        & 12 & 0.213 &  0.214 &  0.214 \\
         \hline
RoBERTa & 0 & 0.810 &  0.770 &  0.770 \\
        & 1 & 0.509 &  0.264 &  0.264 \\
        & 2 & 0.584 &  0.141 &  0.141 \\
        & 3 & 0.607 &  0.152 &  0.152 \\
        & 4 & 0.657 &  0.200 &  0.200 \\
        & 5 & 0.623 &  0.225 &  0.225 \\
        & 6 & 0.641 &  0.242 &  0.242 \\
        & 7 & 0.614 &  0.241 &  0.241 \\
        & 8 & 0.578 &  0.235 &  0.235 \\
        & 9 & 0.591 &  0.270 &  0.270 \\
        & 10 & 0.575 &  0.281 &  0.281 \\
        & 11 & 0.591 &  0.319 &  0.319 \\
        & 12 & 0.566 &  0.301 &  0.301 \\
         \hline
XLNet   & 0 & 0.999 &  0.996 &  0.996 \\
        & 1 & 1.000 &  1.000 &  1.000 \\
        & 2 & 1.000 &  0.987 &  0.987 \\
        & 3 & 0.993 &  0.992 &  0.992 \\
        & 4 & 0.983 &  0.978 &  0.978 \\
        & 5 & 0.903 &  0.896 &  0.896 \\
        & 6 & 0.481 &  0.470 &  0.470 \\
        & 7 & 0.432 &  0.426 &  0.426 \\
        & 8 & 0.235 &  0.236 &  0.236 \\
        & 9 & 0.321 &  0.323 &  0.323 \\
        & 10 & 0.308 &  0.307 &  0.307 \\
        & 11 & 0.124 &  0.150 &  0.150 \\
        & 12 & 0.028 &  0.024 &  0.024 \\
\end{tabular}
\caption{Proportion of variance in cosine similarity \textbf{$r^2$} explained by cosine similarity when the top $k$ dimensions (measured by cosine similarity contribution) are removed. Layer 0 is the static embedding layer.}
\label{table:allvar}
\end{table}

\begin{figure*}[h!]
    \includegraphics[width=\linewidth]{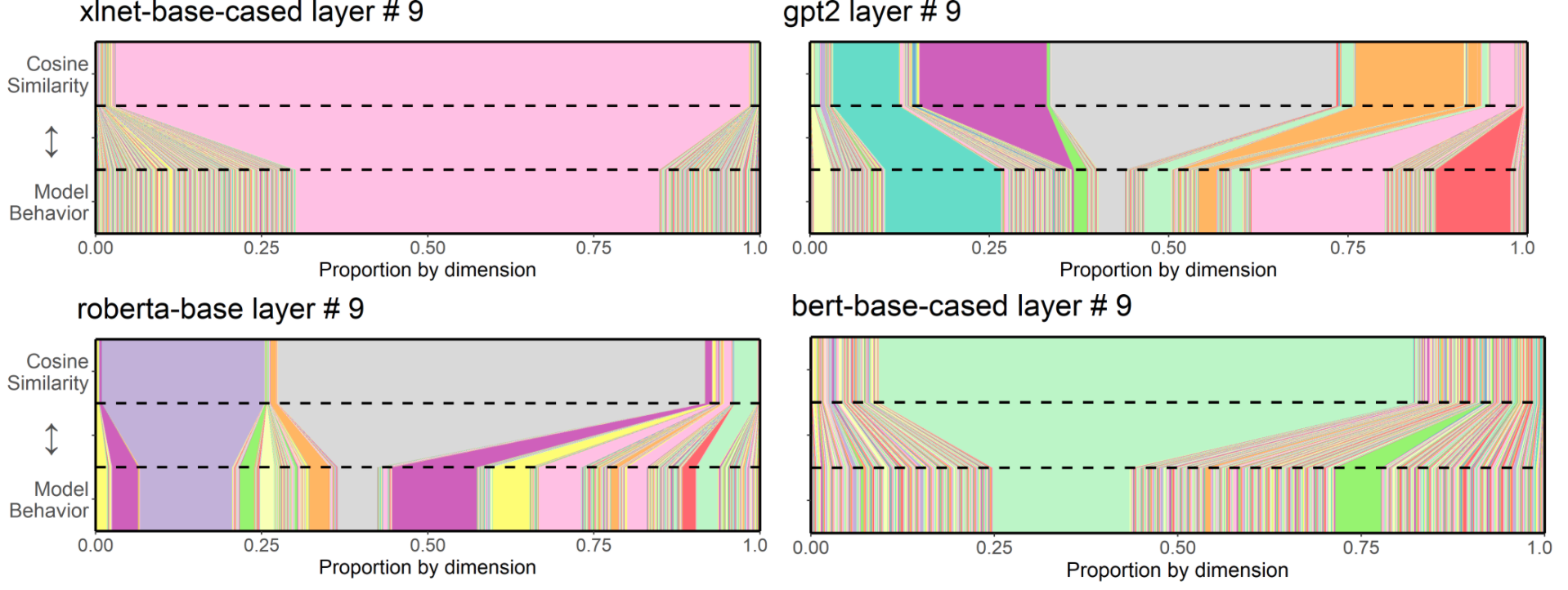}
    \rule{\textwidth}{.5pt}
    \includegraphics[width=\linewidth]{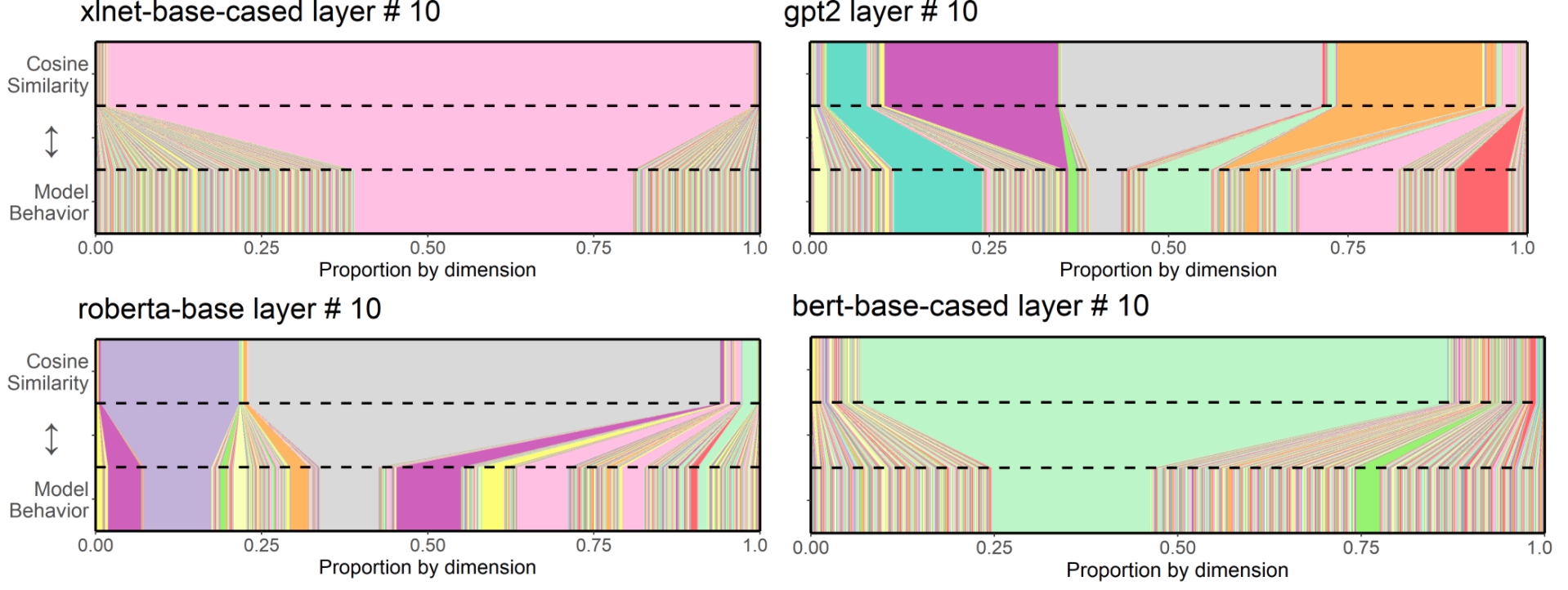}
    \rule{\textwidth}{.5pt}
    \includegraphics[width=\linewidth]{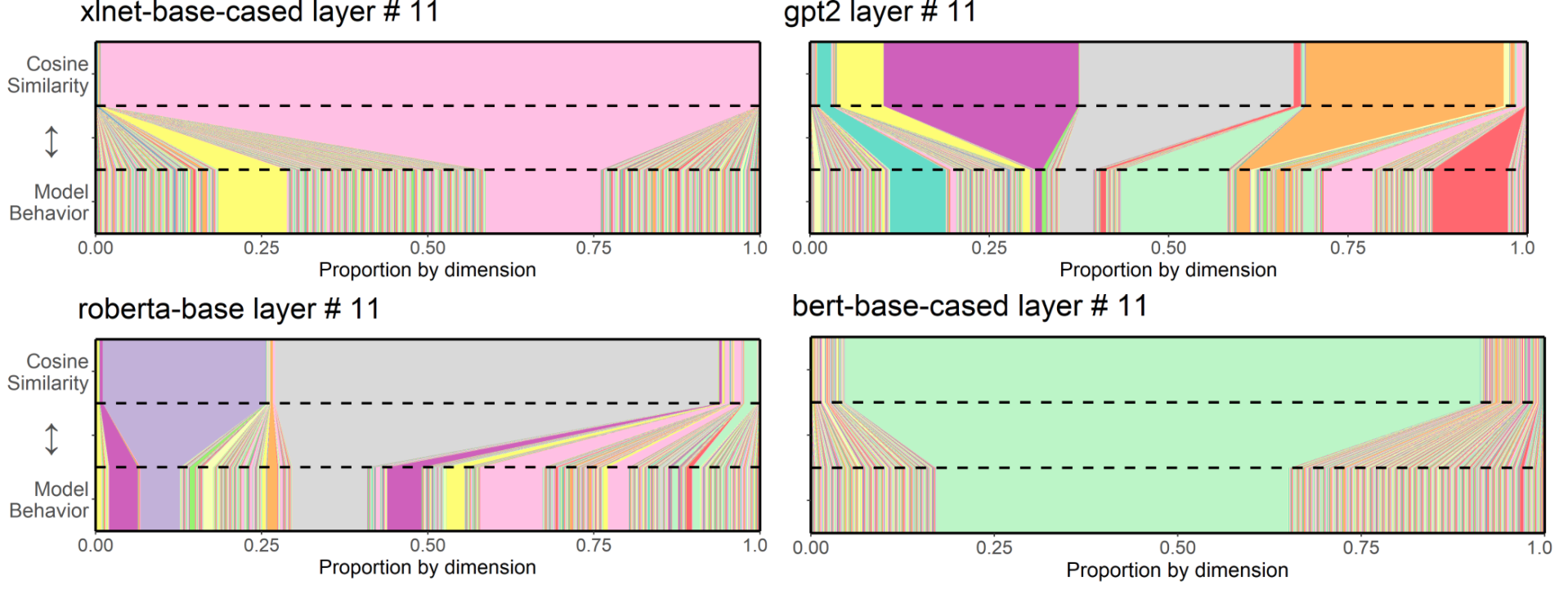}
    \caption{Relative contribution of each dimension to cosine similarity (top) paired with its relative influence on model behavior (bottom) for layers 9-11 of each model.}
    \label{fig:mismatch_9-11}
\end{figure*}

\begin{figure*}[h!]
    \includegraphics[width=\linewidth]{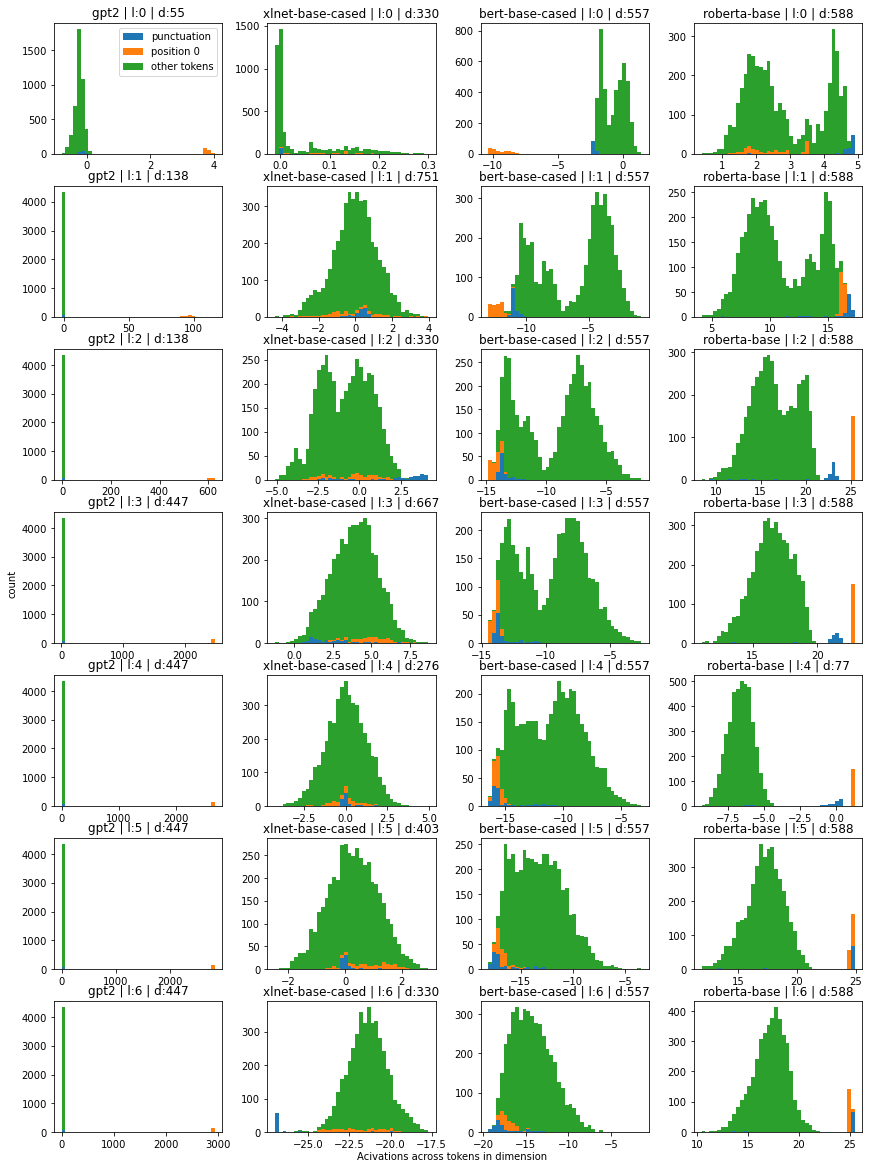}
        \caption{Distribution of activations in the dimension with highest variance in layers 0-6 of each model across a sample of 10k tokens. Each color corresponds to a specific type/position, where the orange distribution is tokens occurring in position zero, the blue distribution is instances of the "." token, and green is all other tokens. In many cases, there are two clear modes in each distribution, where one corresponds to a specific word type or position. Additionally, this behavior tends to persist within the same dimension number across layers, which is facilitated by the residual connections present in each model.}
    \label{fig:rogue_dim_0}
\end{figure*}
\begin{figure*}[h!]
    \includegraphics[width=\linewidth]{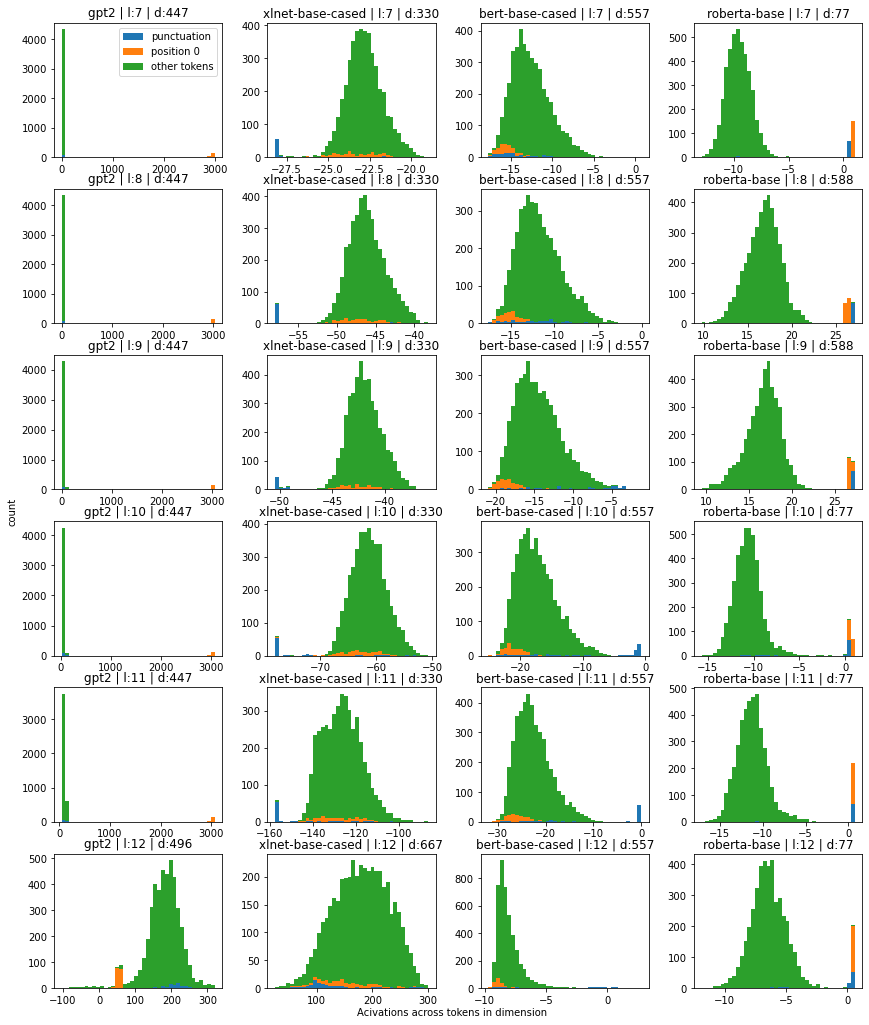}
    \caption{Distribution of activations in the dimension with highest variance in layers 7-12 of each model across a sample of 10k tokens. Each color corresponds to a specific type/position, where the orange distribution is tokens occurring in position zero, the blue distribution is instances of the "." token, and green is all other tokens. In many cases, there are two clear modes in each distribution, where one corresponds to a specific word type or position. Additionally, this behavior tends to persist within the same dimension number across layers, which is facilitated by the residual connections present in each model.}
    \label{fig:rogue_dim_1}
\end{figure*}

\begin{figure*}[h!]
    \includegraphics[width=\linewidth]{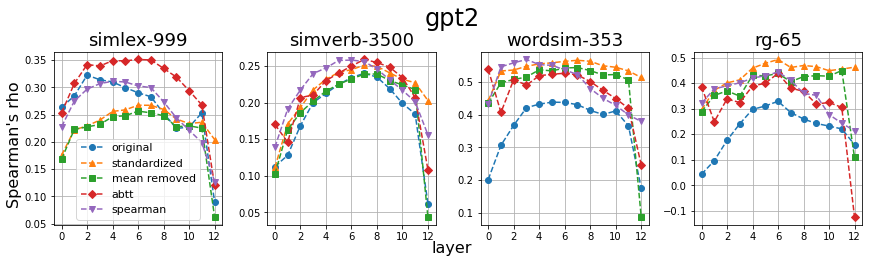}
    \caption{Average correlation (Spearman's $\rho$) with human judgements on each word similarity dataset, with and without postprocessing for GPT-2}
    \label{fig:rep_qualitygpt}
\end{figure*}

\begin{figure*}[h!]
    \includegraphics[width=\linewidth]{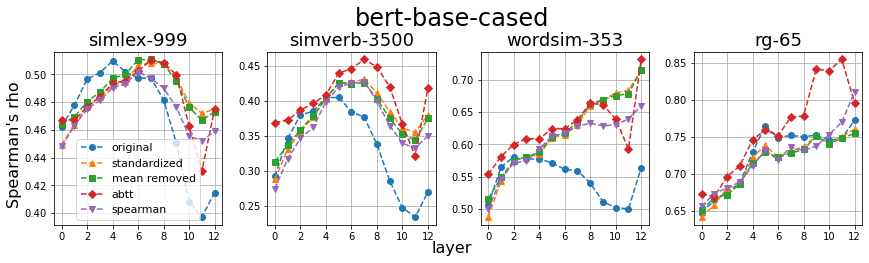}
    \caption{Average correlation (Spearman's $\rho$) with human judgements on each word similarity dataset, with and without postprocessing for BERT}
    \label{fig:rep_qualitybert}
\end{figure*}

\begin{figure*}[h!]
    \includegraphics[width=\linewidth]{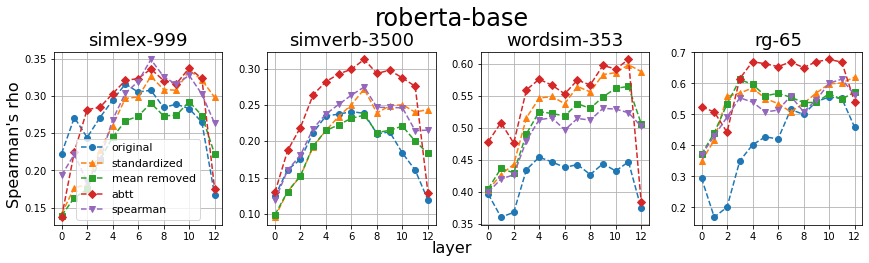}
    \caption{Average correlation (Spearman's $\rho$) with human judgements on each word similarity dataset, with and without postprocessing for RoBERTa}
    \label{fig:rep_qualityroberta}
\end{figure*}

\begin{figure*}[h!]
    \includegraphics[width=\linewidth]{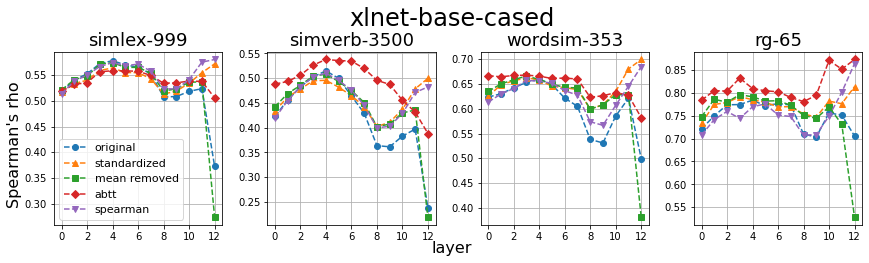}
    \caption{Average correlation (Spearman's $\rho$) with human judgements on each word similarity dataset, with and without postprocessing for XLNet}
    \label{fig:rep_qualityxlnet}
\end{figure*}

\end{document}